\documentclass[conference]{IEEEtran}
\IEEEoverridecommandlockouts
\usepackage{amsmath,amssymb,amsfonts}
\usepackage{algorithmic}
\usepackage{graphicx}
\usepackage{textcomp}
\usepackage{xcolor}
\def\BibTeX{{\rm B\kern-.05em{\sc i\kern-.025em b}\kern-.08em
    T\kern-.1667em\lower.7ex\hbox{E}\kern-.125emX}}

\usepackage{subcaption}
\usepackage{caption}
\usepackage{comment}
\usepackage{multirow}
\usepackage{booktabs}
\usepackage[export]{adjustbox}
\usepackage{dashbox}
\usepackage{enumitem}
\usepackage{tabularx}
\usepackage{array}
\usepackage{arydshln}
\usepackage{balance}
\usepackage[noadjust]{cite}
\usepackage{url}
\usepackage{pgfplots} 
\usetikzlibrary{patterns}
\usepackage{tikz}
\usepackage{textcomp}
\usepackage{hyperref}
\newcommand\copyrighttext{%
\footnotesize \textcopyright 2022 IEEE.
This paper has been accepted for presentation in 23rd IEEE International Conference on Industrial Technology (ICIT22), 22 - 25 August, 2022, Shanghai, China.
Personal use of this material is permitted. Permission from IEEE must be obtained for all other uses, in any current or future media, including reprinting/republishing this material for advertising or promotional purposes, creating new collective works, for resale or redistribution to servers or lists, or reuse of any copyrighted component of this work in other works. DOI: \href{<http://tex.stackexchange.com>}{<DOI No.>}}
\newcommand\copyrightnotice{%
\begin{tikzpicture}[remember picture,overlay]
\node[anchor=south,yshift=10pt] at (current page.south) {\fbox{\parbox{\dimexpr\textwidth-\fboxsep-\fboxrule\relax}{\copyrighttext}}};
\end{tikzpicture}%
}
\begin{document}

\title{Unauthorized Drone Detection: Experiments and Prototypes
\thanks{This publication was made possible by the NPRP award NPRP12S-0313-190348 from the Qatar National Research Fund (a member of The Qatar Foundation). The statements made herein are solely the responsibility of the authors. Qatar National Library funded the publications of this article.}
}

\author{
  \IEEEauthorblockN{
  	Muhammad Asif Khan\IEEEauthorrefmark{1},
    Hamid Menouar\IEEEauthorrefmark{1},
    Osama Muhammad Khalid\IEEEauthorrefmark{1} and
    Adnan Abu-Dayya\IEEEauthorrefmark{1}
  }
  
  \IEEEauthorblockA{
  	Qatar Mobility Innovations Center (QMIC), Qatar University, Doha, Qatar.\IEEEauthorrefmark{1}\\
    Email:
        \{muhammada, hamidm, osamam, adnan\}@qmic.com\IEEEauthorrefmark{1}
  }
  }

\maketitle
\copyrightnotice

\begin{abstract}
The increase in the number of unmanned aerial vehicles a.k.a. drones pose several threats to public privacy, critical infrastructure and cyber security. Hence, detecting unauthorized drones is a significant problem which received attention in the last few years. In this paper, we present our experimental work on three drone detection methods (i.e., acoustic detection, radio frequency (RF) detection, and visual detection) to evaluate their efficacy in both indoor and outdoor environments. Owing to the limitations of these schemes, we present a novel encryption-based drone detection scheme that uses a two-stage verification of the drone's received signal strength indicator (RSSI) and the encryption key generated from the drone's position coordinates to reliably detect an unauthorized drone in the presence of authorized drones.
\end{abstract}


\section{Introduction}

%%%%%%%%  Motivation
Unmanned aerial vehicles (UAVs) a.k.a. drones are becoming a commonplace in many commercial applications such as aerial photography, geological mapping and surveying, traffic and crowd monitoring, search and rescue, emergency response and disaster relief, shipping, crop monitoring and spraying, and emergency communication networking. These widespread applications of drones will lead to an unprecedented growth in the number of commercial drones. According to Federal Aviation Authority (FAA), the number of civilian drones in United States (US) alone is expected to reach 1.66 million by the end of 2023 \cite{faa_report}. The presence of such a huge number of civilian drones in the low-altitude airspace is associated with severe threats such as public privacy invasion, terrorism, and cyber attacks. Thus, to protect people and infrastructure from the illegitimate uses of commercial drones, an efficient stand-alone drone detection system is urgently needed.
\par
Some countries and drone manufacturers have implemented regulatory measures to restrict the illegitimate uses of drones. For instance, some Da-Jiang Innovations (DJI) uses special override commands to restrict drones' entry into no-fly zones. However, these measures do not solve the problem of preventing unauthorized drones operations in non-restricted areas. Furthermore, these restrictions in drones' software can be altered and thus are not sufficient to fully protect NFZs from unauthorized drones' entry.

%%%%%%%%   Techniques
Recently, several drone detection solutions have been proposed by researchers from academia and industry. These systems are typically categorized into four categories based on the underlying detection technology. These include (i) acoustic (sound-based) detection \cite{Shi2020}, (ii) radio frequency (RF) detection \cite{Basak2021}, (iii) radar detection \cite{Morris2021}, and (iv) visual (camera-based) detection \cite{Wei2021, Sie2021, Pavliv2021}.
Acoustic detection uses sound sensors (i.e., microphones) to detect the sound produced by a drone's engine or other auxiliary components such as its propeller blades. The signal is detected by highly sensitive microphones (single or an array of multiple microphones) and processed to extract significant features. The features are then mapped with a database of pre-saved drones' signatures (features indexed by drone's ID) or fed to a machine learning (ML) model to detect and identify the drone. Sound-based detection has a short detection range ($\sim 200$ meter) and is highly susceptible to background noise. However, its advantage is that it does not require visual line-of-sight (LOS) to the target drone.
In RF-based detection, an RF sensor is employed to listen to the control signal between a drone and its ground controller. The RF signals are then processed and fed to a machine learning model to classify the drone.
RF-detection has long detection range (up to several kilometers) and also does not require LOS. However, it fails to detect autonomous drones.
Radar-detection is a relatively more accurate and robust method with proven accuracy in aircraft detection. Radars transmit a high frequency signal and receive the reflected version of it. The doppler shift between the transmitted and reflected signal is used to identify a moving drone.
However, radars fail to detect hovering drones (stationery) and can not differentiate between a drone and a moving bird.
Visual detection uses visual sensors (i.e., daylight and infrared cameras) to monitor the airspace. The video frames are processed using deep learning methods such as convolution neural networks (CNN) to detect the presence of drones.
It is relatively more accurate and efficient, but requires a clear LOS to the target. Furthermore, the accuracy deteriorate in harsh weather such as heavy fog, rain, and dust etc.

\IEEEpubidadjcol  %%% Inserted in the second column to make room for IEEEpubid 

%%%%%%%%%%%  SOTA
\par
The theoretical limitations of drone detection approaches motivates experimental evaluation of these systems. Hence, several recent works dedicate to evaluate the performance of these schemes \cite{khan_sensors_2022}. For instance, \cite{Joel2015, Frank2016, Shi2020} show how acoustic detection can be improved by using different number of microphones. Similarly, the works in \cite{anwar2019, Zahoor2020, Balakin2021} explore feature engineering to process sensors' outputs to improve the detection accuracy. Whereas, \cite{Jeon2017, Shi2020, Kolamunna2021} investigate different machine learning techniques to develop robust system. For RF detection, the research efforts are mainly focused on the advancement and sophistication in the machine learning models to classify the RF signals. Some researchers propose to use simpler ML techniques such as support vector machines (SVM), k-nearest neighbors (kNN), and neural networks (NN) \cite{Ezuma2019}, while others propose deep learning techniques including convolution neural networks (CNN) \cite{Basak2021, Yang2021}, Deep Recurrent Neural Network (DRNN) \cite{Gumaei2021}. Radar-based detection relatively received little attention and is typically focused on Micro-Doppler radars to enhance detection of smaller objects \cite{Bjorklund2018}. Visual detection \cite{White2019, Zheng2021} is gaining popularity due to the growing computer vision community. The recent research in this area advocates CNN-based techniques \cite{Park2017} for higher detection accuracy. While most of the research works attempt to improve the detection accuracy, the problem of unauthorized drone detection in the presence of authorized drones is yet to be investigated.

In this work, we first present our experimental analysis of the three approaches for drone detection i.e., sound-based, RF-based, and camera-based detection. We then propose our prototype system for encryption-based unauthorized drone detection. To the best of our knowledge, the proposed prototype system is novel and is highly effective to detect unauthorised drones in the presence of authorized drones.

%====================================
%% Experiments and Prototypes 
%====================================

\section{Experiments and Prototypes}
\label{sec:implementation}

In this section, we provide details of our experiments and prototypes using three drone detection schemes i.e., acoustic, RF-based, and visual detection. Radar-based detection could not be studied due to regulatory requirements. Additionally, we also provide a novel encryption-based drone detection scheme to cope with the shortcomings of the basic RF detection.

%====================================
%% Accoustic
%====================================
\subsection{Acoustic Detection System}

\begin{figure}[!h]
\centering
\includegraphics[width=0.85\columnwidth]{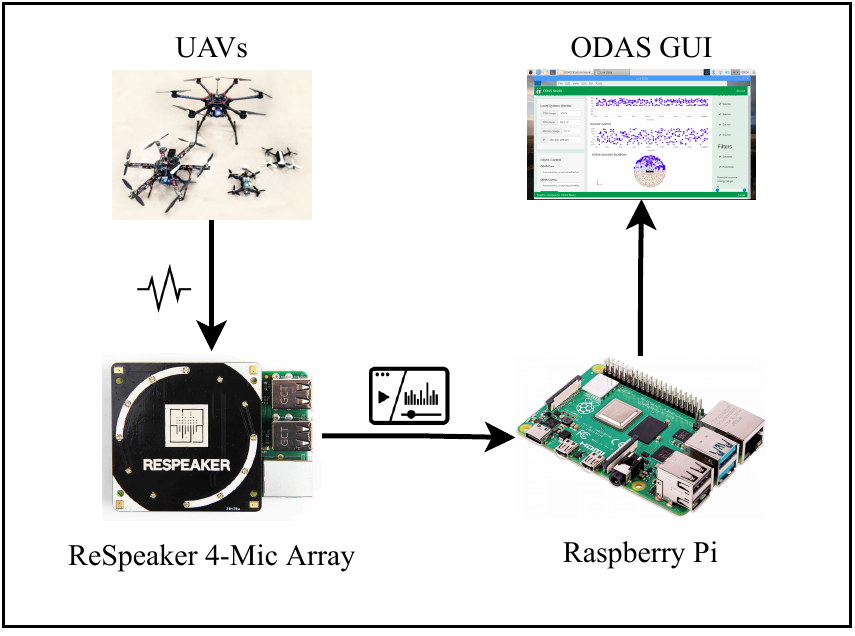}
\caption{Prototype of acoustic drone detection system.}
\label{fig_acoustic_proto}
\end{figure}

%%%% System
The prototype for acoustic detection system is depicted in Fig. \ref{fig_acoustic_proto}. It mainly consists of (i) a sound sensors (i.e., a 4-mic array board, ReSpeaker) to detect drone's sound (ii) a Raspberry Pi processor to process sensor data, and (iii) an Open EmbeddeD Audition System (ODAS) user interface.

%%%%% Figure

The microphones array captures the sound waves and passes it to the Raspberry Pi processor. At Raspberry Pi, the sound waves are processed to detect statistical features. The features are then searched in an existing database of drones' sound signatures to detect the drone's presence and its location information (i.e., source azimuth). These information is then displayed on the ODAS GUI.

%====================================
%% RF
%====================================
\subsection{RF Detection System}

%%% System
Our prototype for RF-based detection using passive detection consists of a universal software radio peripheral (USRP) E312  which is a portable stand-alone software-defined radio (SDR) developed by Ettus research™), an embedded field-programmable gate array (FPGA) along with a dual-core CPU with ARM architecture for accelerating computation, and a GPS device. The USRP comes with a $2\times2$ multiple input-multiple output (MIMO) transceiver with an instantaneous bandwidth of 56 MHz, and a frequency range that varies from 70 MHz up to 6 GHz.

%%% Figure

\begin{figure}[!h]
\centering

%% ======= Method 1 (solid box)
% \fbox{\parbox[c]{0.9\columnwidth}{
% \subcaptionbox{\scriptsize Underground Parking.}{\includegraphics[width=0.8\columnwidth]{figures/rf-1.jpg}}%
% \vspace{2em}
% \subcaptionbox{\scriptsize Indoor Environment.}{\includegraphics[width=0.8\columnwidth]{figures/rf-2.jpg}}%
% }}

%% ======= Method 2 (dashed box)
% \begin{adjustbox}{minipage=0.8\columnwidth, precode=\dbox}
\centering
\subfloat[Underground parking.]{\includegraphics[width=0.8\columnwidth, frame=0.5pt]{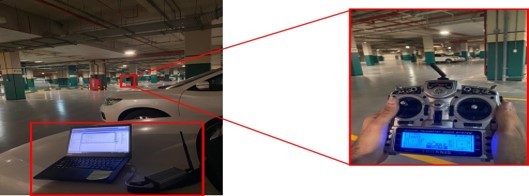}} \\[2em]
\subfloat[Indoor environment.]{\includegraphics[width=0.8\columnwidth, frame=0.5pt]{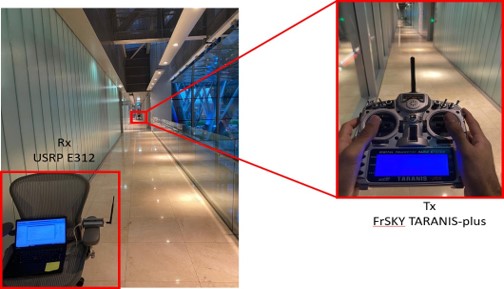}}
% \end{adjustbox}
\caption{RF-based detection prototype.}
\label{fig_rf_proto}
\end{figure}
In our prototype, the USRP has been tuned around 2.4GHz since the sub-band of the controller signals varies from 2.4GHz to 2.47GHz. Then the baseband signal is sampled with a sampling frequency of 20MHz.
The signals were acquired for 100ms and sent through an Ethernet cable to the host computer in the form of frames where each frame is 4k sample long. The sampled signal is then processed at the host computer using a custom-written MATLAB script, where the original signal and its Fourier transform is reconstructed from the IQ complex signal. Then the spectrum of the reconstructed signal is observed in the present and absents of controller signals under different channel condition such as in the present/absent of Wi-Fi signals, Bluetooth signals other types of EM environmental interference that use the same band to establish the communication.
% Fig. \ref{fig_rf_results1} depicts the frequency spectrum in the present and absence of controller signals,
%%% Figure
% \input{figures/fig_rf_results1}

%====================================
%% Visual
%====================================
\subsection{Visual Detection System}

%%% System
The prototype consists of a PTZ (Pan-Tilt-Zoom) camera (model: SD6CE445XA-HNR WizSense series), an NVIDIA Jetson Xavier board to run the CNN model (YOLOv4 \cite{YOLOv4_2020}), and a web interface. The PTZ camera is installed on a box facing upward to cover the sky view. The camera is programmed to continuously scan for the presence of drone operating in two operating modes i.e., (i) user-defined, and (ii) step scanning. In user-defined mode, the user set the camera zoom, starting angle, and final angle of the camera. The camera then continuously scan the specified field of view (FOV). In the step mode, horizontal FOV is divided into steps instead of a continuous range. The camera changes its FOV by changing its FOV by one step, stops for a second, capture a frame, and move to the next step. The experimental set up in depicted in Fig. \ref{fig_rf_proto} showing multiple scenarios i.e., in an underground parking garage, inside building corridors, and outdoor environment.

%====================================
%% Encryption-based
%====================================
\subsection{Encryption-based Detection System}
\begin{figure}[!h]
\centering

\subcaptionbox{The BLE Broadcast module.}{\includegraphics[width=0.9\columnwidth]{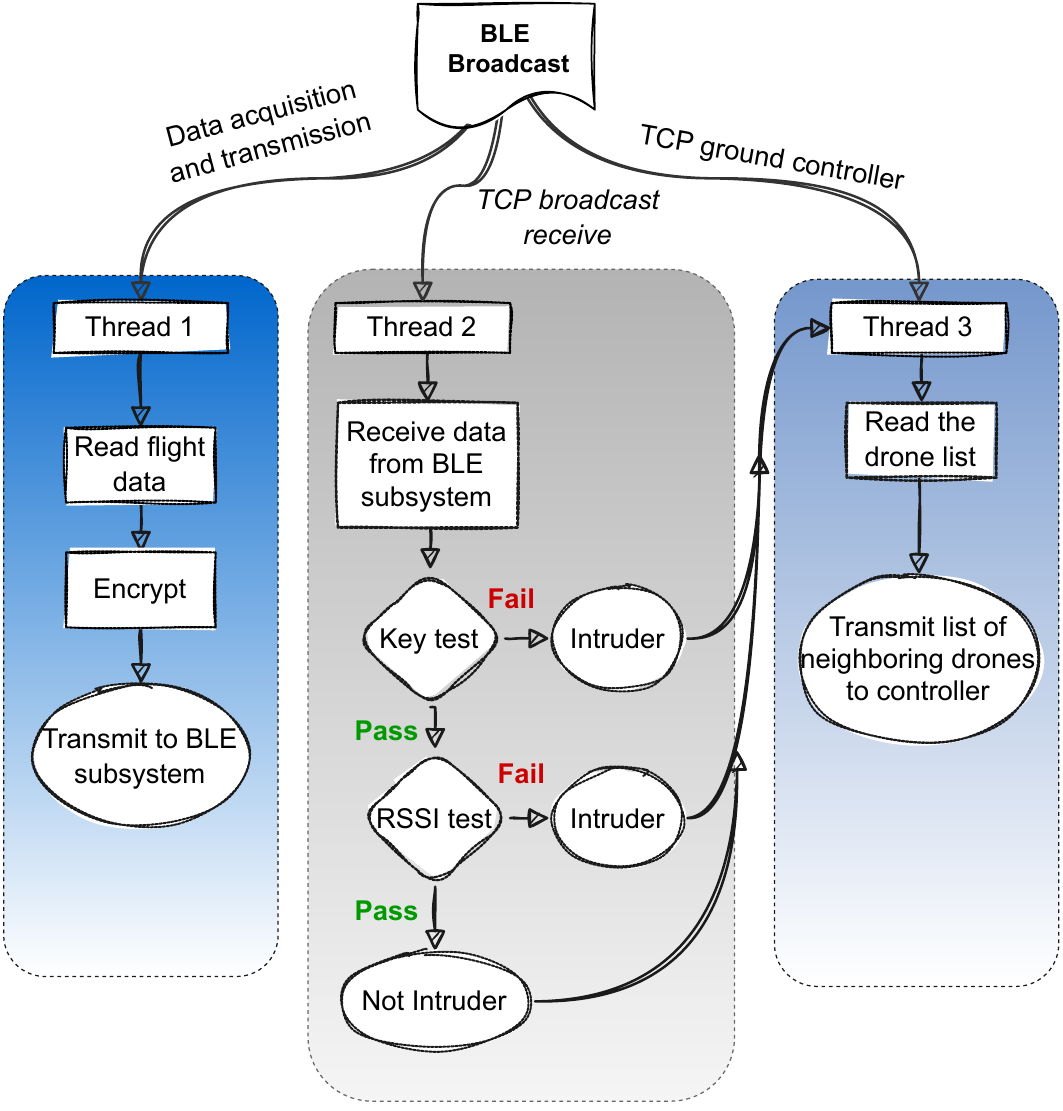}}
\hspace{1em}
\subcaptionbox{The BLE Subsystem module.}{\includegraphics[width=0.85\columnwidth]{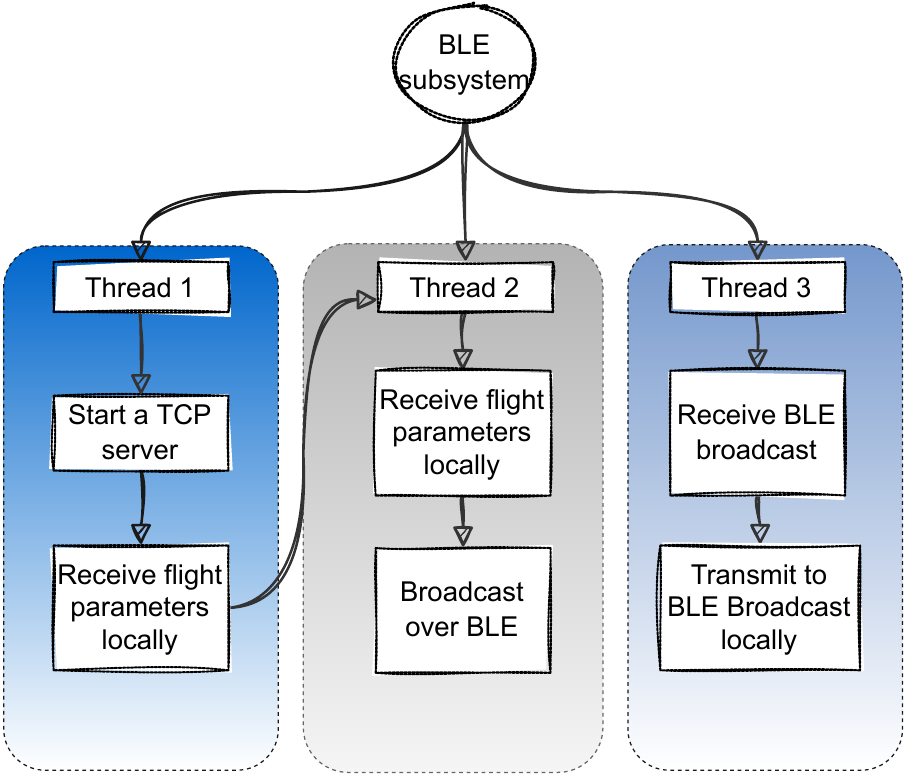}}
%%%%%%% SVG
% \subcaptionbox{The BLE Broadcast module.}{\includesvg[width=0.85\columnwidth](figures/script11.svg)}
% \hspace{1em}
% \subcaptionbox{The BLE Subsystem module.}{\includesvg[width=0.8\columnwidth](figures/script22.svg)}

\caption{The proposed encryption-based detection scheme.}
\label{fig_encryption_scheme}
\end{figure}

The encryption-based method is proposed to detect an unauthorized drone in a swarm of authorized drones. Our prototype system consists of two drones and a ground controller (Holybro Pixhawk Mini). The drones are equipped with Raspberry Pi 4 Model B processors, while the controller is equipped with NVIDIA Jetson AGX Xavier Developer Kit and is integrated with the web-portal user interface. The functional diagram for our encryption-based detection system is exhibited in Fig. \ref{fig_encryption_scheme}. The proposed scheme makes use of the RF-based detection with additional encryption method to increase the reliability of the detection.

%%%%%   Figure

Each drone runs two scripts i.e., \textit{BLE broadcast} and \textit{BLE sub-system}. The BLE broadcast is responsible for acquiring all drone's information from the controller. It consists of three \textit{threads}. The first thread continuously reads the drone’s data, generates the verification key, and sends to the BLE subsystem. The second thread receives the drone's flight data, the encryption key, and RSSI for each drone from the BLE subsystem. The key and RSSI are both verified. If any of the two tests for key and RSSI matching fails, the drone's list is updated by flagging the drone as \textit{unauthorized}. The updated list is shared with the third thread in the BLE broadcast. The list updates every 0.1 seconds. The third thread reads the updated list and transmit it to the controller via Wi-Fi link. The encryption key is generated from drone's ID and its GPS coordinates as follow.
\begin{equation*}
    KEY = \{(5 \times ID) + (4 \times LAT) + (3\times LON) + (2\times ALT)\} \%256
    \label{eq:key}
\end{equation*}
The modulus 256 is used to generate a key of size 8 bits which can be easily packed in the BLE advertisement packet. The BLE subsystem is responsible to broadcast flight data from all drones along with their RSSI values and the encryption key over standard Bluetooth advertisements. It also consists of three threads. The first thread receive extracted data from drone’s flight controller via the BLE Broadcast script locally over TCP. The second thread receives the flight data from the first thread and formats the data into BLE packets. The third thread receives the BLE advertisement packets, unpacks, and transmits them locally to the BLE Broadcast script.

%====================================
%% Results
%====================================
\section{Results} \label{sec:results}
\begin{figure}[!h]
    \centering
    \includegraphics[width=0.99\columnwidth]{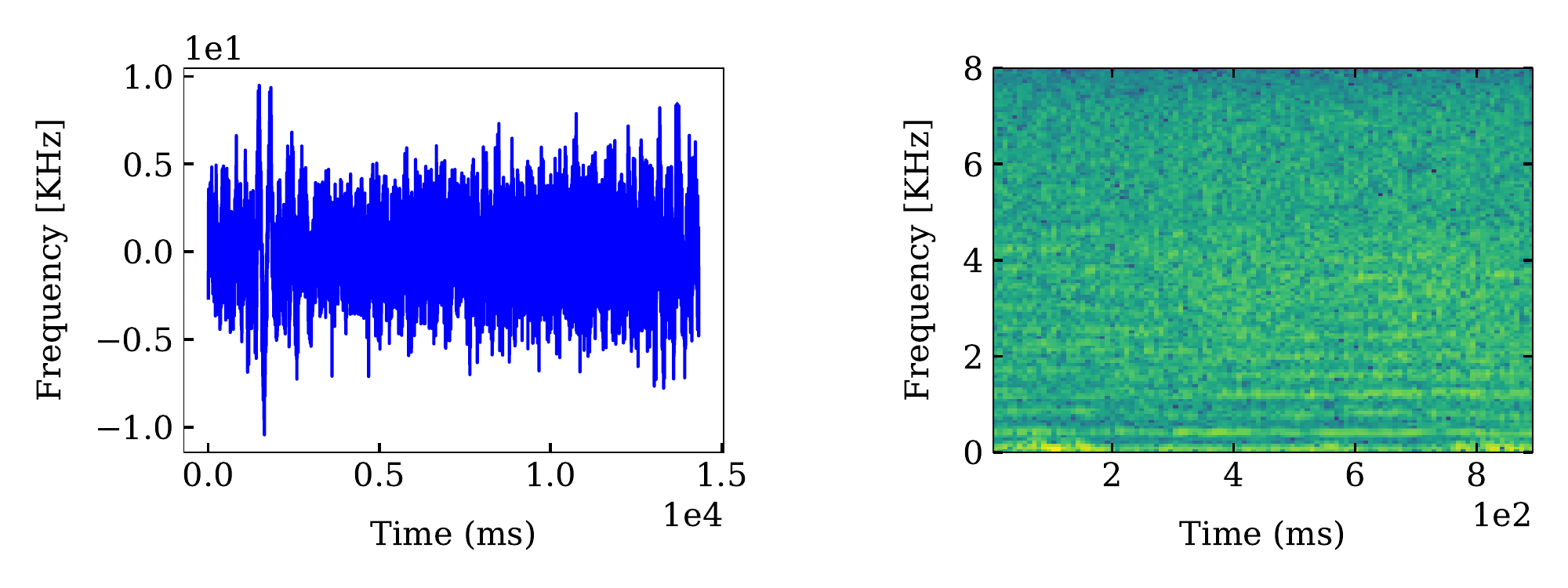}
    \caption{Acoustic signature of the flying drone.}
    \label{fig_acoustic_signature}
\end{figure}

%%%%% Acoustic
We evaluated the performance of each detection scheme. In acoustic detection, sound signatures of each drone were analyzed using a spectogram. A sample sound signature and its spectogram is presented in Fig. \ref{fig_acoustic_signature}.

\begin{figure}[!h]
\centering
\includegraphics[frame=1pt, width=7cm, height=5cm]{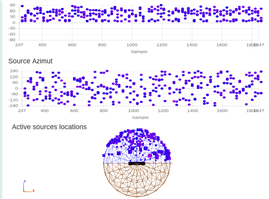}
\caption{A snapshot of ODAS GUI showing drone's estimated locations.}
\label{fig_acoustic_results}
\end{figure}

We used a direction of arrival (DOA) extraction algorithm which outputs up to 4 source angels. Fig. \ref{fig_acoustic_results} shows the location estimates from acoustic signatures. We observed random errors in the drone's location due to significant background noise in outdoor environment. This affirms that the use of larger arrays of highly sensitive microphones to improve the performance of acoustic detection \cite{Frank2016}.

\begin{figure}[!h]
\centering
\subcaptionbox{ROC Curve.}{\includegraphics[width=0.95\columnwidth]{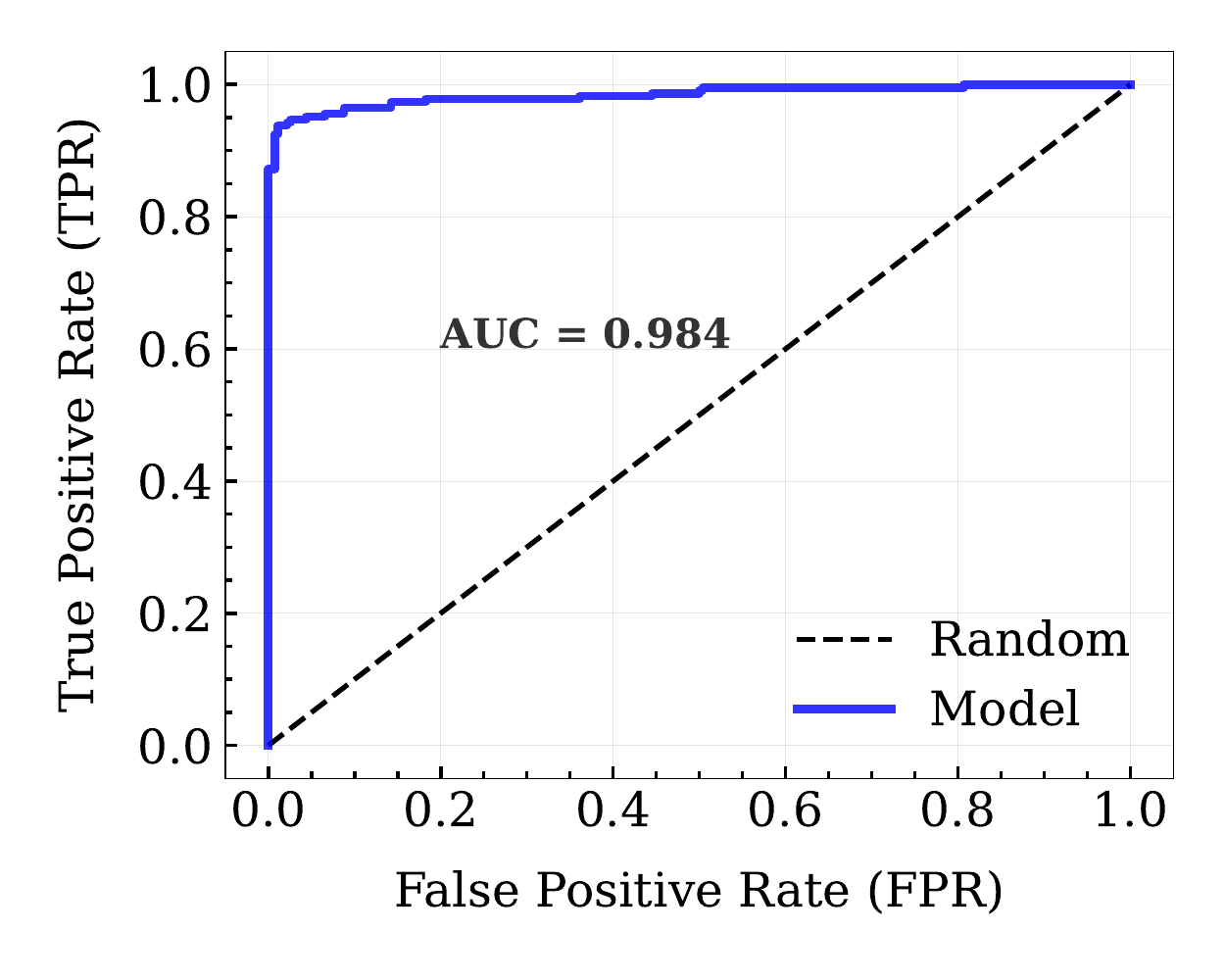}}%rf-roc.png  roc_curve.pdf
\vspace{0.2em} \\
\subcaptionbox{Precision-Recall Curve.}{\includegraphics[width=0.95\columnwidth]{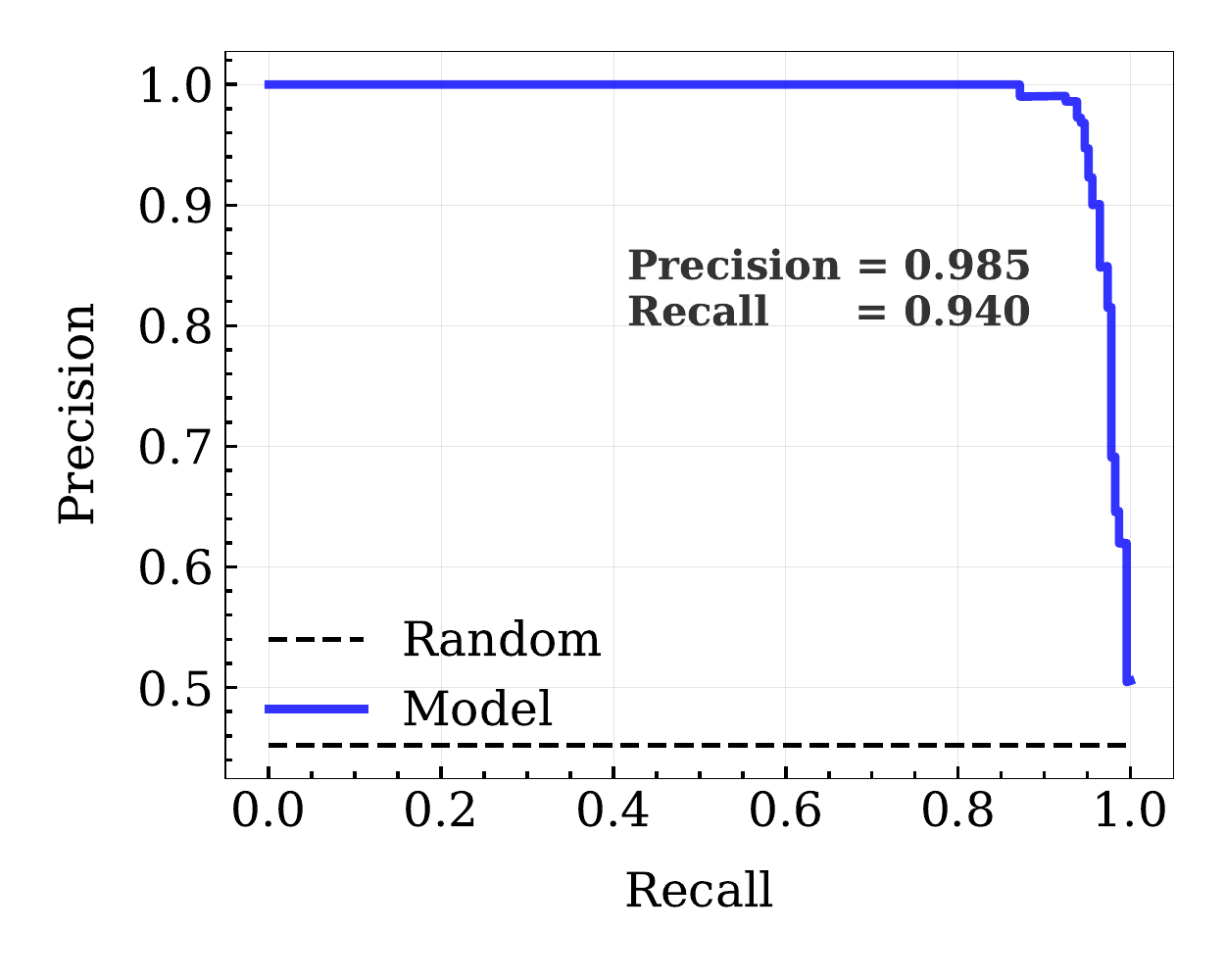}}%rf-precision.png  precision_recall_curve.pdf
\caption{RF-based detection performance analysis.}
\label{fig_rf_results}
\end{figure}

%%%%%% RF
To evaluate RF-based detection system, several tests are conducted in multiple scenarios as depicted in Fig. \ref{fig_rf_proto}. Raw data ($\sim 1000$ samples) was collected to form a balanced binary labeled dataset and then to construct the receiver operating characteristic curve (ROC) as depicted in Fig. \ref{fig_rf_results}.

%%% Figure

It can be observed in Fig. \ref{fig_rf_results} that the area under the ROC curve (AUC) reaches to almost $0.98$. The optimal operating point is acquired from the ROC curve which corresponds to the threshold value of $74.02$, precision $0.985$ and recall $0.94$.

%%%%% Table

\begin{table}[!h]
\caption{System performance under different threshold values.}
\label{tab_rf_perf}
\centering
\renewcommand{\arraystretch}{1.2}
\begin{tabular}{|c|m{1.5cm}|m{1.5cm}|m{1.5cm}|}
% \toprule
\hline
Threshold &Accuracy &Precision &Recall \\ 
% \midrule
\hline
60 &0.9281 &0.8927 &0.9673 \\ 

65 &0.9459 &0.9376 &0.9510 \\ 

70 &0.9596 &0.9687 &0.9469 \\ 

75 &0.9656 &0.9851 &0.9429 \\

80 &0.9587 &0.9912 &0.9224 \\ 

85 &0.9537 &0.9978 &0.9061 \\ 
% \bottomrule
\hline
\end{tabular}
\end{table}

%%% Comments
The results in Table \ref{tab_rf_perf} presents the the recall, precision, and accuracy of the proposed system under different threshold values. These values clearly show the reliability and the robustness of the developed system. However, we expect that the performance of this system can potentially degrade when adopted in large-scaled problems and other more dynamic environments.
One possible way to reduce false alarms is to introduce \textit{two-samples, one-alarm policy}, in which a drone will only be declared in case of two consecutive positive detections. This also eliminates the need to adjust the system threshold. The downside is that it compromises the real-time performance of the system due to doubling of the processing.
% A possible extension is to employ RF detection as part of a hybrid system. Usually when we reduce the specified threshold, the recall of the system increases which is a desirable improvement. However, this improvement will come at the cost of the system’s precision, but this can be tolerated since the system is only a part of the hybrid system and it is only used to trigger other detection mechanism to acknowledge or deny the alarm before declaring the existence of a drone.

\begin{figure}[!h]
\centering
\includegraphics[width=0.9\columnwidth, frame=3pt]{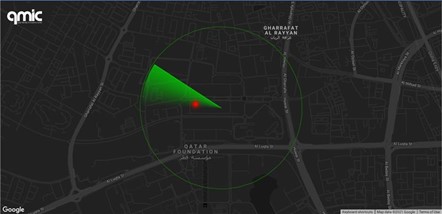}
\caption{Graphical interface of camera-based detection prototype system: An unauthorized drone is detected. The drone is shown as red circle while the camera's FOV is displayed as green beam. The beam provides the estimated location and direction of the detected drone.}
\label{fig_visual_proto}
\end{figure}

%%%%%%%%%%%%%%%%%%  Visual
In visual detection, we used You Only Look Once (YOLOv4) deep learning model to train a model on a two-class dataset containing $944$ drone images and $270$ helicopter images collected from several public datasets. We also acquired more drone images using the PTZ camera and labelled these new images. We further trained the YOLOv4 model using transfer learning on the server. The system achieved $90\%$ prediction accuracy on the test data (unseen). The trained model is then converted into TensorRT model and deployed over NVIDIA Jetson Xavier. When a drone is detected in a video frame captured by the camera, the system automatically estimates the location of the drone based on the direction of the camera. We developed an algorithm to automatically adjust the camera direction to track a moving drone. The algorithm uses the detection box dimensions, and updates the camera direction based on the box dimensions relative to the frame. The web-interface showing a detected drone in red color whereas the camera's FOV as a green beam is depicted in Fig. \ref{fig_visual_proto}. The width of the beam indicates the current zoom level of the camera i.e., beam width is automatically adjusted as inversely proportional to the zoom level.

%%%%% Figure

%%%% Comments
It is worthy to note that the use of a single camera can provide limited coverage of the region of interest at a time. To monitor $360^\circ{}$ area, multiple cameras configured in an array at the same or different locations can be used. Camera-based detection also has a limited range and can't detect drone that are too far due to limited optical zoom capability. Visual detection also requires LOS to the target, we suggest to integrate acoustic or RF detection as a first stage to provide an estimated location of the drone to the camera to rotate towards the target drone.

\begin{figure}[!h]
    \centering
    \includegraphics[width=0.9\columnwidth]{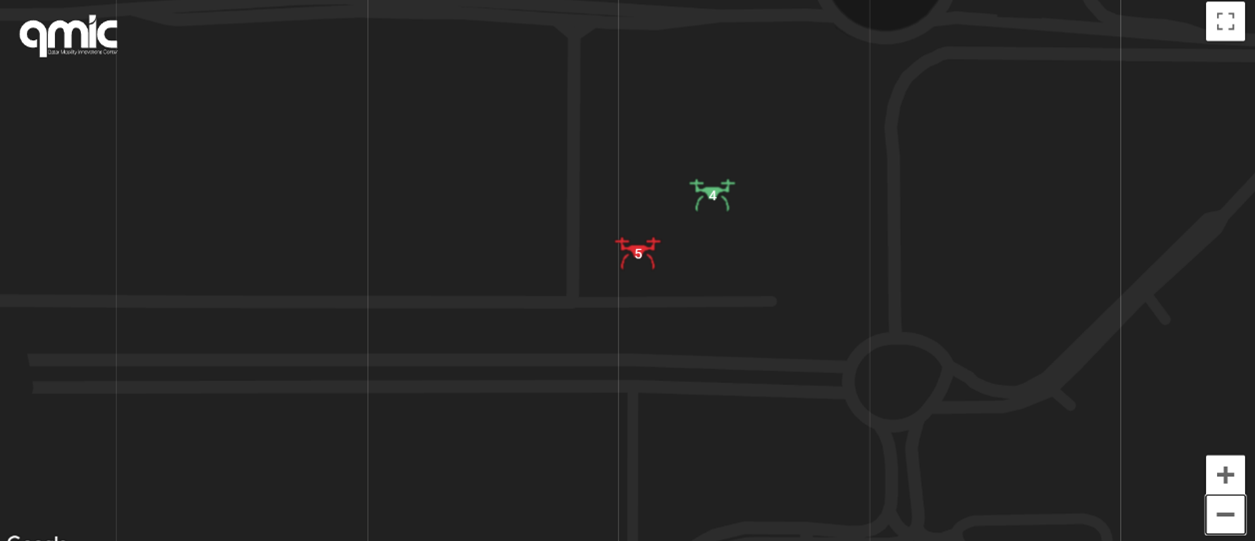}
    \caption{Graphical interface of encryption-based detection prototype system: An intruder drone (red) located close to a a non-intruder drone (green) is accurately detected and localized on the map.}
    \label{fig_encryption_gui}
\end{figure}

%%%%%%%%%%%%%%%%%  Encryption-based
Lastly, we evaluated our encryption-based scheme on real drones for accuracy. The proposed scheme produced highly reliable results in different set ups. Fig. \ref{fig_encryption_gui} shows the intruder drone detected using the proposed prototype.

%====================================
%% Conclusion
%====================================
\section{Conclusion and Future Work}
In this paper, we present our preliminary experiments and prototypes for three drone detection systems i.e., acoustic detection, RF-based detection, and visual detection respectively. We present only candidate results for each method (due to space limitation). However, our detailed investigation shows serious limitations of each method in different scenarios. We propose a novel encryption-based unauthorized drone detection in the presence of other authorized drones. The prototype is experimentally validated which accurately detects an unauthorized drone in the presence of an authorized drone. As future work, we suggest other encryption-based schemes (e.g., \cite{jiang_2022}) as well as sensor fusion to overcome the limitations of individual drone detection system. This may include a two-stage hybrid system using sensor fusion e.g., hybrid RF and visual sensors system. In such a system, the RF sensor provides an early detection of the presence of a drone at a larger distance and provides an estimated location /direction of the remote drone to the camera which detects, identify, and tracks the drone more accurately (second stage). Furthermore, the visual detection approach rely on the availability of large datasets of drone's images, whereas the existence dataset only contains images of few drone models. The existing visual detection schemes can be further improved and tested on large datasets of containing signatures of a variety of drone models with different prospective and scale captures in diverse environmental conditions.

% \section*{Acknowledgment}
% % \scriptsize
% This work was supported in part by the Qatar National Research Fund (a member of Qatar Foundation) through the NPRP Grant under Grant NPRP12S-0313-190348. The statements made herein are solely the responsibility of the authors.

% put at least one blank line to end the scriptsize paragraph and
% then revert back to normalsize.
\normalsize

\balance
%%%%%%%%%%%%%%%%%%%%%%%%%%%%%%%%%%%%%%%%%%%
%%%%%    References
%%%%%%%%%%%%%%%%%%%%%%%%%%%%%%%%%%%%%%%%%%%

\bibliographystyle{ieeetr} %IEEEtran
\bibliography{biblio}

\begin{thebibliography}{10}

\bibitem{faa_report}
``Faa airspace forecast - fiscal year 2019-2039.''
  \url{https://www.faa.gov/data_research/aviation/aerospace_forecasts/media/FY2019-39_FAA_Aerospace_Forecast.pdf}.
\newblock Accessed: 2021-12-15.

\bibitem{Shi2020}
Z.~Shi, X.~Chang, C.~Yang, Z.~Wu, and J.~Wu, ``An acoustic-based surveillance
  system for amateur drones detection and localization,'' {\em IEEE
  Transactions on Vehicular Technology}, vol.~69, no.~3, pp.~2731--2739, 2020.

\bibitem{Basak2021}
S.~Basak, S.~Rajendran, S.~Pollin, and B.~Scheers, ``Drone classification from
  rf fingerprints using deep residual nets,'' in {\em 2021 International
  Conference on COMmunication Systems NETworkS (COMSNETS)}, pp.~548--555, 2021.

\bibitem{Morris2021}
P.~J.~B. Morris and K.~V.~S. Hari, ``Detection and localization of unmanned
  aircraft systems using millimeter-wave automotive radar sensors,'' {\em IEEE
  Sensors Letters}, vol.~5, no.~6, pp.~1--4, 2021.

\bibitem{Wei2021}
D.~T. Wei~Xun, Y.~L. Lim, and S.~Srigrarom, ``Drone detection using yolov3 with
  transfer learning on nvidia jetson tx2,'' in {\em 2021 Second International
  Symposium on Instrumentation, Control, Artificial Intelligence, and Robotics
  (ICA-SYMP)}, pp.~1--6, 2021.

\bibitem{Sie2021}
N.~J. Sie, S.~Srigrarom, and S.~Huang, ``Field test validations of vision-based
  multi-camera multi-drone tracking and 3d localizing with concurrent camera
  pose estimation,'' in {\em 2021 6th International Conference on Control and
  Robotics Engineering (ICCRE)}, pp.~139--144, 2021.

\bibitem{Pavliv2021}
M.~Pavliv, F.~Schiano, C.~Reardon, D.~Floreano, and G.~Loianno, ``Tracking and
  relative localization of drone swarms with a vision-based headset,'' {\em
  IEEE Robotics and Automation Letters}, vol.~6, no.~2, pp.~1455--1462, 2021.

\bibitem{khan_sensors_2022}
M.~A. Khan, H.~Menouar, A.~Eldeeb, A.~Abu-Dayya, and F.~D. Salim, ``On the
  detection of unauthorized drones - techniques and future perspectives: A
  review,'' {\em IEEE Sensors Journal}, pp.~1--1, 2022.

\bibitem{Joel2015}
{Joël Busset and Florian Perrodin and Peter Wellig and Beat Ott and Kurt
  Heutschi and Torben Rühl and Thomas Nussbaumer}, ``{Detection and tracking
  of drones using advanced acoustic cameras},'' in {\em Unmanned/Unattended
  Sensors and Sensor Networks XI; and Advanced Free-Space Optical Communication
  Techniques and Applications} (E.~M. Carapezza, P.~G. Datskos, C.~Tsamis,
  L.~Laycock, and H.~J. White, eds.), vol.~9647, pp.~53 -- 60, International
  Society for Optics and Photonics, SPIE, 2015.

\bibitem{Frank2016}
{Frank Christnacher and Sébastien Hengy and Martin Laurenzis and Alexis
  Matwyschuk and Pierre Naz and Stéphane Schertzer and Gwenael Schmitt},
  ``{Optical and acoustical UAV detection},'' in {\em Electro-Optical Remote
  Sensing X} (G.~Kamerman and O.~Steinvall, eds.), vol.~9988, pp.~83 -- 95,
  International Society for Optics and Photonics, SPIE, 2016.

\bibitem{anwar2019}
M.~Z. Anwar, Z.~Kaleem, and A.~Jamalipour, ``Machine learning inspired
  sound-based amateur drone detection for public safety applications,'' {\em
  IEEE Transactions on Vehicular Technology}, vol.~68, no.~3, pp.~2526--2534,
  2019.

\bibitem{Zahoor2020}
Z.~Uddin, M.~Altaf, M.~Bilal, L.~Nkenyereye, and A.~K. Bashir, ``Amateur drones
  detection: A machine learning approach utilizing the acoustic signals in the
  presence of strong interference,'' {\em Computer Communications}, vol.~154,
  pp.~236--245, 2020.

\bibitem{Balakin2021}
M.~Balakin, A.~Dvorak, and D.~Kurylev, ``Real-time drone detection and
  recognition by acoustic fingerprint,'' in {\em 2021 5th Scientific School
  Dynamics of Complex Networks and their Applications (DCNA)}, pp.~44--45,
  2021.

\bibitem{Jeon2017}
S.~Jeon, J.~Shin, Y.-J. Lee, W.-H. Kim, Y.~Kwon, and H.-Y. Yang, ``Empirical
  study of drone sound detection in real-life environment with deep neural
  networks,'' {\em 2017 25th European Signal Processing Conference (EUSIPCO)},
  pp.~1858--1862, 2017.

\bibitem{Kolamunna2021}
H.~Kolamunna, T.~Dahanayaka, J.~Li, S.~Seneviratne, K.~Thilakaratne, A.~Y.
  Zomaya, and A.~Seneviratne, ``Droneprint: Acoustic signatures for open-set
  drone detection and identification with online data,'' {\em Proc. ACM
  Interact. Mob. Wearable Ubiquitous Technol.}, vol.~5, mar 2021.

\bibitem{Ezuma2019}
M.~Ezuma, F.~Erden, C.~K. Anjinappa, O.~Ozdemir, and I.~Guvenc, ``Micro-uav
  detection and classification from rf fingerprints using machine learning
  techniques,'' in {\em 2019 IEEE Aerospace Conference}, pp.~1--13, 2019.

\bibitem{Yang2021}
S.~Yang, Y.~Luo, W.~Miao, C.~Ge, W.~Sun, and C.~Luo, ``Rf signal-based uav
  detection and mode classification: A joint feature engineering generator and
  multi-channel deep neural network approach,'' {\em Entropy}, vol.~23, no.~12,
  2021.

\bibitem{Gumaei2021}
A.~Gumaei, M.~Al-Rakhami, M.~M. Hassan, P.~Pace, G.~Alai, K.~Lin, and
  G.~Fortino, ``Deep learning and blockchain with edge computing for 5g-enabled
  drone identification and flight mode detection,'' {\em IEEE Network},
  vol.~35, no.~1, pp.~94--100, 2021.

\bibitem{Bjorklund2018}
S.~Bjorklund, ``Target detection and classification of small drones by boosting
  on radar micro-doppler,'' in {\em 2018 15th European Radar Conference
  (EuRAD)}, pp.~182--185, 2018.

\bibitem{White2019}
I.~White, D.~K. Borah, and W.~Tang, ``Robust optical spatial localization using
  a single image sensor,'' {\em IEEE Sensors Letters}, vol.~3, no.~6, pp.~1--4,
  2019.

\bibitem{Zheng2021}
Y.~Zheng, Z.~Chen, D.~Lv, Z.~Li, Z.~Lan, and S.~Zhao, ``Air-to-air visual
  detection of micro-uavs: An experimental evaluation of deep learning,'' {\em
  IEEE Robotics and Automation Letters}, vol.~6, no.~2, pp.~1020--1027, 2021.

\bibitem{Park2017}
J.~Park, D.~H. Kim, Y.~S. Shin, and S.-h. Lee, ``A comparison of convolutional
  object detectors for real-time drone tracking using a ptz camera,'' in {\em
  2017 17th International Conference on Control, Automation and Systems
  (ICCAS)}, pp.~696--699, 2017.

\bibitem{YOLOv4_2020}
A.~Bochkovskiy, C.-Y. Wang, and H.-Y.~M. Liao, ``Yolov4: Optimal speed and
  accuracy of object detection,'' {\em ArXiv}, vol.~abs/2004.10934, 2020.

\bibitem{jiang_2022}
Y.~Jiang, S.~Wu, H.~Yang, H.~Luo, Z.~Chen, S.~Yin, and O.~Kaynak, ``Secure data
  transmission and trustworthiness judgement approaches against cyber-physical
  attacks in an integrated data-driven framework,'' {\em IEEE Transactions on
  Systems, Man, and Cybernetics: Systems}, pp.~1--11, 2022.

\end{thebibliography}

\end{document}